\documentclass[letterpaper, 10 pt, conference]{ieeeconf}
\IEEEoverridecommandlockouts 
\usepackage{cite}
\usepackage{amsmath,amssymb,amsfonts}
\usepackage{algorithmic}
\usepackage{textcomp}

\usepackage[utf8]{inputenc}
\usepackage[english]{babel}

\usepackage{mathtools}
\usepackage{amsfonts}
\usepackage[hidelinks]{hyperref}
\usepackage[acronym,abbreviations]{glossaries-extra}
\usepackage{glossaries-prefix}
\usepackage[capitalise]{cleveref}
\usepackage{textcomp}
\usepackage{float}
\usepackage{array}
\usepackage{import}
\usepackage{MnSymbol}
\usepackage{algorithm2e}
\usepackage{bbm}
\usepackage{xparse}
\usepackage{pgfplots}
\usepackage{pgfplotstable}
\usepackage{wrapfig}
\usepackage{subcaption}
\usepgfplotslibrary{fillbetween}
\usepackage{xcolor}
\usepackage{autonum}
\usepackage[font=small]{caption}
\usepackage{microtype}
\usepackage{makecell}
\usepackage{tikz}
\usetikzlibrary{arrows,automata}
\usetikzlibrary{external}
\usepackage{graphicx}
\PassOptionsToPackage{pdftex}{graphicx}

\RestyleAlgo{ruled}

\crefname{equation}{Eq.}{Eqs.}
\crefname{section}{Sec.}{Secs.}
\crefname{figure}{Fig.}{Figs.}

\definecolor{pltBlue}{HTML}{1f77b4}
\definecolor{pltOrange}{HTML}{ff7f0e}
\definecolor{pltGreen}{HTML}{2ca02c}
\definecolor{pltPurple}{HTML}{9467bd}
\definecolor{pltPink}{HTML}{e377c2}
\definecolor{pltLightBlue}{HTML}{17becf}

\pgfplotsset{
    every axis plot/.append style={line width=0.8pt},
    every axis plot post/.append style={
        every mark/.append style={mark=none}
    }
}

\pgfplotsset{std legend/.style={
  legend image code/.code={\fill[#1] (0cm,-0.07cm) rectangle (0.6cm,0.07cm);}
  }
}

\NewDocumentCommand\plotstd{mmmmmmoO{1}O{1}}{%
    \pgfplotstableread[col sep=comma]{#1}\datatable
    \IfNoValueTF{#7}{}{%
        \addplot[smooth, opacity=0, name path=A, forget plot] table [x expr=#9 * \thisrow{#5}, y expr=#3 * (\thisrow{#6} - #8 * \thisrow{#7})] {\datatable};
        \addplot[smooth, opacity=0, name path=B, forget plot] table [x expr=#9 * \thisrow{#5}, y expr=#3 * (\thisrow{#6} + #8 * \thisrow{#7})] {\datatable};
        \addplot[#2, opacity=0, fill opacity=0.4]  fill between [of=A and B];
        \addlegendentry{$\pm$ std};
    }
    \addplot+[color=#2, smooth, solid] table [x expr=#9 * \thisrow{#5}, y expr=#3 * \thisrow{#6}] {\datatable};
    \addlegendentry{#4};
}

\NewDocumentCommand\plotstdcom{mmmmmmoO{1}O{1}}{%
    \pgfplotstableread[col sep=comma]{#1}\datatable
    \IfNoValueTF{#7}{}{%
        \addplot[smooth, opacity=0, name path=A, forget plot] table [x expr=#9 * \thisrow{#5}, y expr=#3 * (\thisrow{#6} - #8 * \thisrow{#7})] {\datatable};
        \addplot[smooth, opacity=0, name path=B, forget plot] table [x expr=#9 * \thisrow{#5}, y expr=#3 * (\thisrow{#6} + #8 * \thisrow{#7})] {\datatable};
        \addplot[#2, opacity=0, fill opacity=0.4, forget plot]  fill between [of=A and B];
    }
    \addplot+[color=#2, smooth, solid] table [x expr=#9 * \thisrow{#5}, y expr=#3 * \thisrow{#6}] {\datatable};
    \addlegendentry{#4};
}

\pgfplotsset{rewplot/.style={
    align=center,
    legend style={nodes={scale=0.6, transform shape}},
    xlabel = {\footnotesize Training epochs}, 
    ylabel near ticks,
    xlabel near ticks,
    grid=both,
    width=1.1\textwidth,
    height=4.5cm,
    reverse legend,
    legend pos=south east,
    legend cell align={left}
}}

\DeclareMathOperator*{\argmax}{arg\,max}
\DeclareMathOperator*{\Ex}{\mathbb{E}}
\DeclareMathOperator*{\Var}{\mathrm{Var}}

\newcommand{\rawModel}{\emph{Inception-RI}}
\newcommand{\ptModel}{\emph{Inception-PT}}
\newcommand{\smallModel}{\emph{Inception-S}}

\newcommand{\newdataset}{\emph{Active Clothing Perception Dataset}}

\setabbreviationstyle[acronym]{long-short}

\newcommand{\anaacr}[3]{%
    \newacronym[prefixfirst={a\ },prefix={an\ }]{#1}{#2}{#3}%
}

\newacronym{ep}{EP}{Exploratory Procedure}
\anaacr{mlbp}{MLBP}{Multi-Local Binary Pattern}
\newacronym{ann}{NN}{Neural Network}
\newacronym{cnn}{CNN}{Convolutional Neural Network}
\newacronym{uq}{UQ}{Uncertainty Quantification}
\newacronym{rnn}{RNN}{Recurrent Neural Network}

\makeglossaries
\glsdisablehyper

\begin{document}
 
\title{What Matters for Active Texture Recognition\\With Vision-Based Tactile Sensors\\}

\author{
    Alina Böhm\textsuperscript{1}, Tim Schneider\textsuperscript{1}, Boris Belousov\textsuperscript{2}, Alap Kshirsagar\textsuperscript{1},\\Lisa Lin\textsuperscript{3}, Katja Doerschner\textsuperscript{3}, Knut Drewing\textsuperscript{3}, Constantin A. Rothkopf\textsuperscript{4,5}, Jan Peters\textsuperscript{1,2,4,5}
\thanks{\textsuperscript{1}Intelligent Autonomous Systems Lab, Department of Computer Science, TU Darmstadt, Germany, {\tt\small tim.schneider1@tu-darmstadt.de}}
\thanks{\textsuperscript{2}German Research Center for AI (DFKI) \url{http://dfki.de/sairol}}
\thanks{\textsuperscript{3}Department of Psychology, University of Giessen, Germany}
\thanks{\textsuperscript{4}Centre for Cognitive Science, Technical University of Darmstadt}
\thanks{\textsuperscript{5}Hessian Center for Artificial Intelligence (Hessian.AI), Darmstadt}
\thanks{We thank Hessisches Ministerium für Wissenschaft \& Kunst for the DFKI grant and ``The Adaptive Mind'' grant.}
}

\maketitle

\begin{abstract}
This paper explores active sensing strategies that employ vision-based tactile sensors for robotic perception and classification of fabric textures.
We formalize the active sampling problem in the context of tactile fabric recognition and provide an implementation of information-theoretic exploration strategies based on minimizing predictive entropy and variance of probabilistic models. 
Through ablation studies and human experiments, we investigate which components are crucial for quick and reliable texture recognition.
Along with the active sampling strategies, we evaluate neural network architectures, representations of uncertainty, influence of data augmentation, and dataset variability.
By evaluating our method on a previously published Active
Clothing Perception Dataset and on a real robotic system, we establish that the choice of the active exploration strategy has only a minor influence on the recognition accuracy, whereas data augmentation and dropout rate play a significantly larger role.
In a comparison study, while humans achieve~$66.9\%$ recognition accuracy, our best approach reaches~$90.0\%$ in under~$5$ touches, highlighting that vision-based tactile sensors are highly effective for fabric texture recognition.
\end{abstract}

\section{Introduction}

\global\csname @topnum\endcsname 0
\global\csname @botnum\endcsname 0
\label{sec:introduction}

Touch is a crucial sensing modality that helps humans perceive object properties and perform dexterous manipulation tasks.
Without tactile feedback, even simple tasks such as lighting a match become harder to perform~\cite{johansson2009coding,dang2014stable}.
Therefore, incorporating tactile sensing into robotics is an important step towards making robots more versatile and dexterous~\cite{lederman1993extracting}.
In this paper, we focus on the problem of tactile perception of object properties, and in particular, on the recognition of fabric textures.
Various applications, such as laundry separation and fabric recycling, waste sorting, and material handling, can benefit from rapid texture classification, as discussed in~\cite{yuan2018active}.

\begin{figure}[t]
    \centering
    \begin{subfigure}{0.15\linewidth}
        \centering
        \includegraphics[height=2.1cm, width=\linewidth]{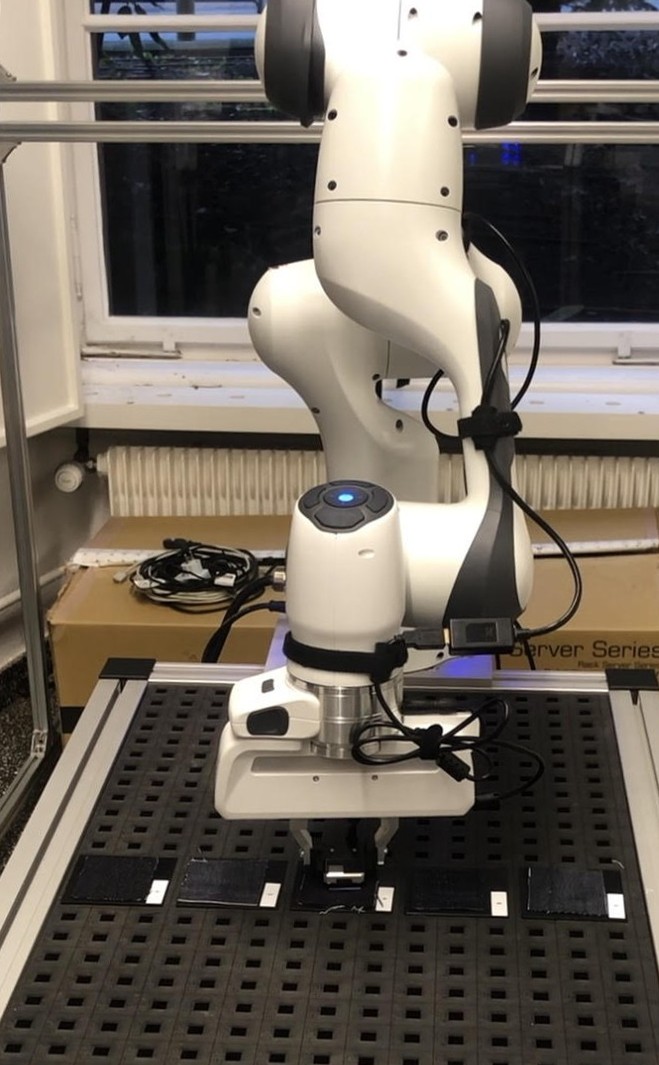}
        \caption{}
        \label{subfig:robot_sensor}
    \end{subfigure}
    \begin{subfigure}{0.245\linewidth}
        \centering
        \includegraphics[height=2.1cm, width=\linewidth]{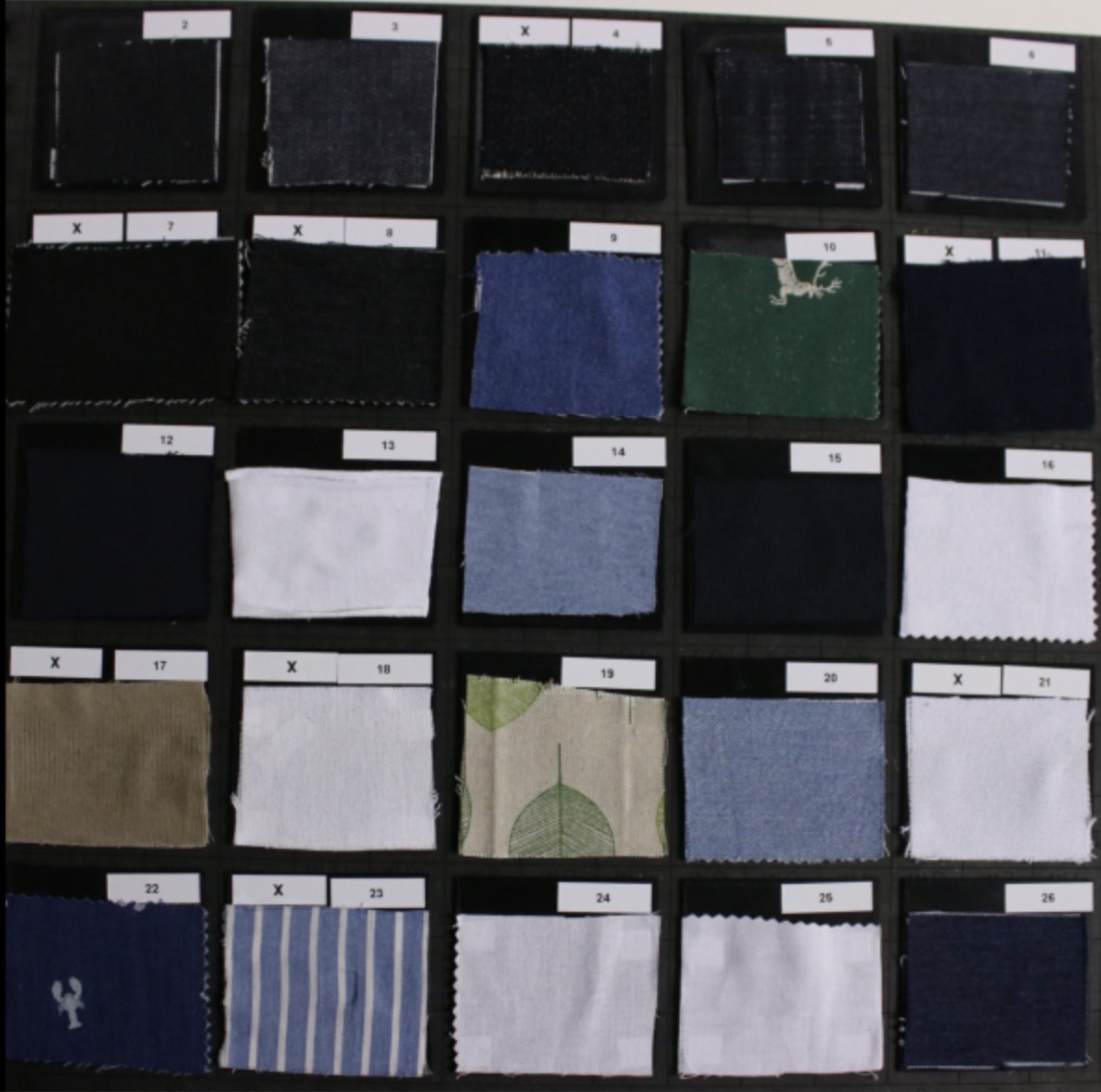}
        \caption{}
        \label{subfig:single_fabric}
    \end{subfigure}
    \begin{subfigure}{0.31\linewidth}
        \centering
        \includegraphics[height=2.1cm, width=\linewidth]{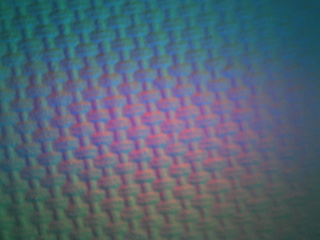}
        \caption{}
        \label{subfig:gelsightimg}
    \end{subfigure}
    \begin{subfigure}{0.25\linewidth}
        \centering
        \includegraphics[height=2.1cm, width=\linewidth]{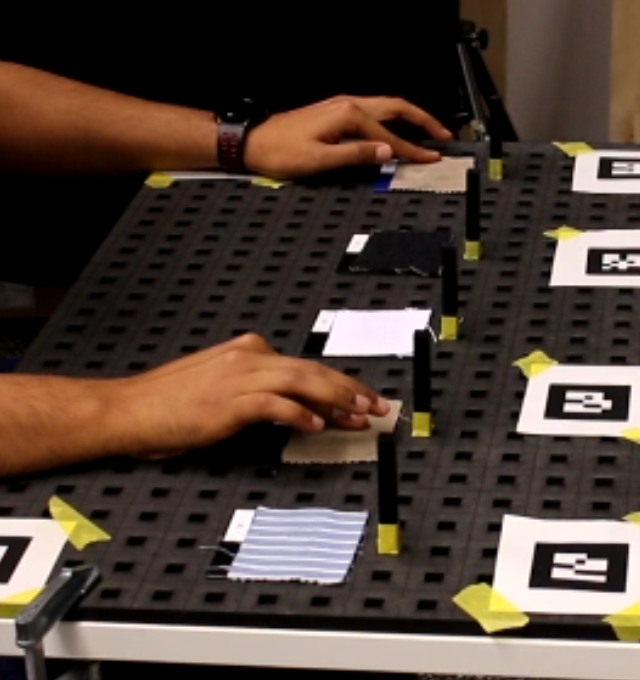}
        \caption{}
        \label{subfig:human_setup}
    \end{subfigure}
    \caption{%
    The texture recognition task requires identifying a given fabric among four comparison samples:
    (\subref{subfig:robot_sensor}) robot arm exploring sample fabrics;
    (\subref{subfig:single_fabric}) dataset of $25$ fabrics;
    (\subref{subfig:gelsightimg}) example tactile image;
    (\subref{subfig:human_setup}) human participant using index fingers to compare fabric samples.
    }
    \label{fig:fabrics}
    \vspace{-1em}
\end{figure}

Classification of fabrics has been tackled with different types of sensors using both supervised and active methods.
\emph{Vibration/force-based tactile sensors}
of different types---such as the iCub sensors~\cite{taunyazov2019towards, taunyazov2020fast, gao2021explainability}, BioTac sensors~\cite{taunyazov2020fast, gao2021explainability}, and custom-designed ones~\cite{lima2020dynamic, huang2021texture, wang2022fabric}---have been used for supervised texture classification, using spiking neural networks~\cite{taunyazov2020fast}, modified RNNs~\cite{gao2021explainability}, and k-NN classifiers~\cite{wang2022fabric}.
All these methods rely on high-frequency temporal data, requiring RNNs or spatio-temporal subsampling to keep the input dimensionality low.
In contrast, \emph{vision-based tactile sensors} provide high-resolution data but at a lower rate, thereby requiring less history as input.
Supervised classification of fabrics was successfully showcased using GelSight heightmap patterns~\cite{li2013sensing} and more advanced spatio-temporal attention features~\cite{cao2020spatio}.
Furthermore, \emph{active sampling methods} have been developed for GelSight to `actively' collect the data~\cite{yuan2018active}, in the sense of repeating touches until a `good' tactile image is obtained, or for material roughness classification~\cite{amini2020uncertainty}, where predictions on image patches were weighted by the output variance of a Bayesian CNN to improve the overall label prediction accuracy.

In this paper, we are tackling the problem of \emph{tactile active texture recognition} (see~\cref{fig:fabrics}).
With no pre-training, a robot is given a `reference' texture and asked to identify it among four comparison textures using as few touches as possible.
This problem setup models applications where the robot needs to quickly identify an object provided only a few touches.
Unlike~\cite{yuan2018active}, we do not want to pre-train on a large dataset but rather quickly adapt on-the-fly, and we do not aim to `classify' but only `recognize' fabrics.
In contrast to~\cite{amini2020uncertainty}, we do not use uncertainty for label prediction but rather for action selection, choosing which fabric to touch next.
This setup also allows us to compare robot vs. human tactile exploration strategies.

In the next sections, we formalize the tactile active texture recognition problem, present a general Bayesian decision-theoretic framework for action selection, describe our implementation which leverages probabilistic NNs for uncertainty quantification, and provide extensive empirical studies and analysis of different components of the algorithm, including the comparison to human exploration strategies and ablations on two datasets and experiments on a real robot.

\section{Problem Setup and Task Formalization}
\label{sec:methods}
We investigate sample-efficient texture recognition using vision-based tactile sensors such as GelSight Mini~\cite{gelsightmini}, Digit~\cite{lambeta2020digit}, or FingerVision~\cite{yamaguchi2019recent}.
In this paper, we focus on the GelSight Mini sensor, as it provides high-resolution, high-quality images, independent of the external lighting.
The sensor is held by a Franka Panda~\cite{franka} robotic arm (see~\cref{fig:fabrics}) and pressed against pieces of fabrics on plastic platforms at predefined locations with randomized amounts of pressure and rotation around the vertical axis to provide more variability in the data.
The leftmost platform holds the \emph{reference texture}, while the remaining four platforms hold randomly chosen \emph{comparison textures}, one of which is equal to the reference.

The agent's goal is to identify the reference among the comparison textures using as few touches as possible.
Crucially, the agent has no prior knowledge of the textures, and therefore has to learn about them within one \emph{trial}, i.e., one fixed selection of five fabrics in a particular order.
One trial consists of multiple touches and ends after a predefined number of touches in our robot experiments or once the participant has made a decision in our human study.
The \emph{action} of the agent is the high-level choice which platform to approach next (the low-level robot control is handled by a Cartesian position controller).
We call each step of this action-observation loop a \emph{round}, and we start counting rounds after each object has been touched once, i.e., if the process has terminated after one round, it means the agent has touched all four comparison fabrics and the reference fabric once and then did just one additional touch.
Thus, one trial consist of several rounds (up to $20$).
We perform multiple trials with different textures, and multiple \emph{runs} for each trial to reduce the statistical error.

To provide textures for our experiments, we created a dataset of 25 denim and cotton fabrics, chosen to be particularly hard to distinguish by touch, as confirmed by our human study in~\cref{subsec:human_experiment}.
For each fabric, we collected $200$ \emph{samples} with randomly perturbed positions and rotations around the vertical axis.
A sample of this dataset can be seen in~\cref{subfig:gelsightimg}.
Our complete dataset is available online.\footnote{Our Tactile Active Recognition of Textures (TART) Dataset can be downloaded at \url{https://drive.google.com/drive/folders/1S_2PLKV-Ap2tifV1gvMfCYjoaNyQaF9z?usp=sharing}}

\section{Tactile Active Texture Recognition Method}
Consider one round of the agent's decision making.
Having touched each of the five platforms one or more times, the agent needs to make a decision which platform to touch next.
The Bayesian approach to this problem is to build a probabilistic model and to choose the action that provides the most information to support the final decision (i.e., the decision which fabric is identical to the reference)~\cite{settles2009active}.
To implement this approach, we specify the model, describe how it is updated using the new data, and define the \emph{acquisition function}, i.e., the action selection strategy.

\subsection{Probabilistic Model Specification}
\label{subsec:prob_classifier}
As the output of the GelSight Mini sensor is a $320\times240$ RGB image, one either needs to manually extract features or employ a CNN.
In~\cite{li2013sensing}, heightmap patterns were used, but with the advent of deep learning, automated feature extraction is prevalent.
Therefore, in this paper, we employ a CNN with dropout layers to implement a probabilistic classifier~\cite{abdar2021review}.
Dropout has been shown to provide a viable approach for uncertainty quantification with neural networks~\cite{gal2016dropout}.
Our experiments with an ensemble of CNNs have shown similar performance to dropout, albeit at a higher computational cost.

\begin{figure}[t]
    \centering
    \begin{subfigure}{0.48\linewidth}
    \includegraphics[height=1.2cm, 
    width=\linewidth]{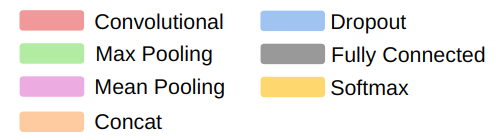}
        \vspace{-1em}
        \label{fig:small}
    \end{subfigure}
    \begin{subfigure}{0.5\linewidth}
        \includegraphics[width=\linewidth, height=0.9cm]{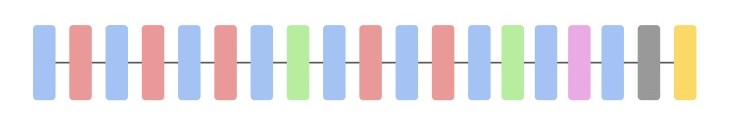}
        \vspace{-1.5em}
        \caption{\smallModel~with added dropout}
        \label{fig:smalldrop}
    \end{subfigure}
    \begin{subfigure}{0.95\linewidth}
        \includegraphics[width=\linewidth, height=2.4cm
        ]{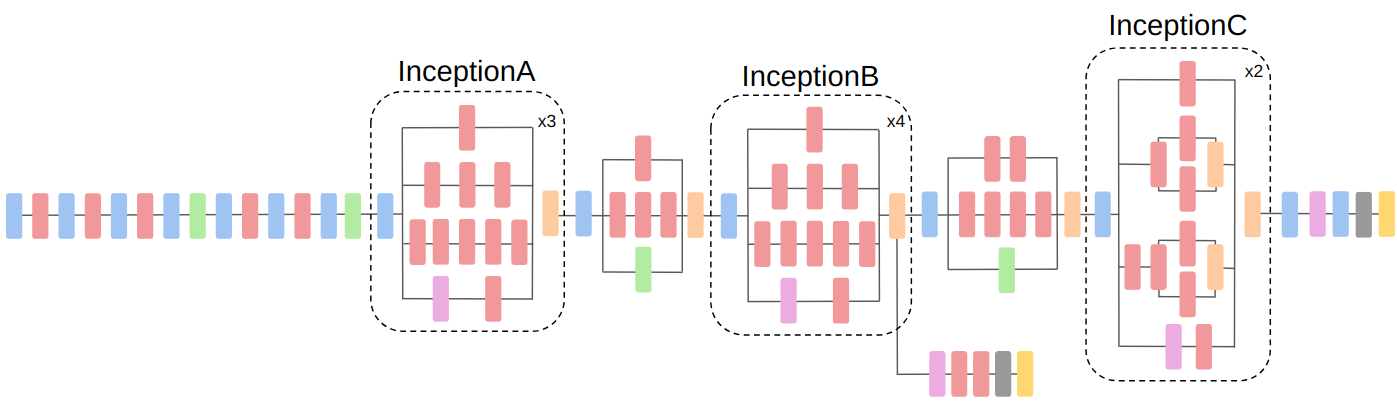}
        \vspace{-1em}
        \caption{Inception-v3 with added dropout}
        \label{fig:incdrop}
    \end{subfigure}
    \caption{%
        The considered architectures of the probabilistic classifier: Inception-v3 and small Inception-v3 (\smallModel) with dropout.
    }
    \label{fig:architectures}
    \vspace{-1em}
\end{figure}

We consider three CNN variants: i) Inception-v3~\cite{szegedy2016rethinking} pre-trained on the ImageNet~\cite{deng2009imagenet} (\ptModel); ii) randomly initialized Inception-v3 (\rawModel); iii) small unpretrained version of Inception-v3 (\smallModel), which drops all the layers after the first \emph{InceptionA} block and before the last \emph{InceptionC} block (see Fig.~\ref{fig:architectures}).
Considering these network variants allows us to evaluate the effects of pretraining and the network depth.
We furthermore add dropout layers and evaluate different dropout rates in our ablation studies.

\subsection{Model Update}
Once a new tactile image is obtained, the model needs to be updated to incorporate the new evidence.
As is common in deep learning, we employ \emph{data augmentation}~\cite{shorten2019survey}, by generating $10$ randomly rotated versions of the same tactile image.
Using all the samples collected during the current trial, we retrain the probabilistic NN classifier:
the samples of the comparison textures serve as inputs and the respective platform positions serve as labels.
The output of the classifier is a probability distribution $p_{\theta}(i|o)$ over the platform labels $i \in \{1, 2, 3, 4\}$, given an image $o$ and the model parameters~$\theta$.

Hence, the model learns to map texture samples to platform labels.
When queried with the reference texture (unseen during training), the model outputs a `probability distribution' over the labels.
To obtain a more robust estimate, we apply the model to $10$ randomly rotated copies of the reference image and average the probabilities
\begin{equation}
    i^\ast = \argmax_i \frac{1}{n_{\text{ref}}} \sum_{k = 1}^{n_{\text{ref}}} p_{\theta}(i|o^{\text{ref}}_k)
\end{equation}
where $n_{\text{ref}}$ is the number of samples of the reference texture and $o^{\text{ref}}_k$ is the $k$-th reference sample. This response is correct if the platform with the same fabric as the reference is predicted.

\subsection{Active Sample Selection Strategy}
\label{subsec:strategy_intro}
The decision which platform to explore next is made based on the model uncertainty.
As described in \cref{subsec:prob_classifier}, we add dropout layers to Inception-v3 (see~\cref{fig:architectures}) to model the epistemic uncertainty~\cite{gal2016dropout}.
By querying the dropout network with the same input multiple times, we obtain different output samples and can gauge the uncertainty by their distribution.

We compare four sample selection strategies: \emph{Random}, \emph{Variance}, \emph{Entropy}, and \emph{You Only Touch Once (YOTO)}.\\
i) \emph{Random} strategy is a naive non-active baseline that selects the next texture to touch according to a uniform distribution.
ii)~\emph{Variance} strategy selects the platform for which the variance of the class probability predictions is the highest
\begin{equation}
    i_{\text{next}} 
    = \argmax_i 
    \frac{1}{n_{\text{ref}}} \sum_{k = 1}^{n_{\text{ref}}} 
    \Var_{m \sim p(m)}[ p_{\theta}(i|o^{\text{ref}}_k, m) ]
\end{equation}
where $n_{\text{ref}}$ is the number of collected samples of the reference object and $p(m)$ is the distribution of the dropout masks.
iii)~\emph{Entropy} strategy selects the platform that contributes the most to the class distribution entropy for the reference object
\begin{equation}
    i_{\text{next}} 
    = \argmax_i  
     \frac{1}{n_{\text{ref}}} \sum_{k = 1}^{n_{\text{ref}}} 
    \Ex_{m \sim p(m)} [
    -p_i^k \ln p_i^k
    ]
\end{equation}
where $n_{\text{ref}}$ and $p(m)$ defined as before and $p_i^k \coloneqq p_{\theta}(i|o^{\text{ref}}_k, m)$.
iv)~\emph{You Only Touch Once (YOTO)} is a trivial baseline that makes a decision immediately after the initial five touches, i.e., each object touched once.
This baseline provides a reference to quantify the `value' of the actively gathered data.

\section{Experimental Results}
\label{sec:results}

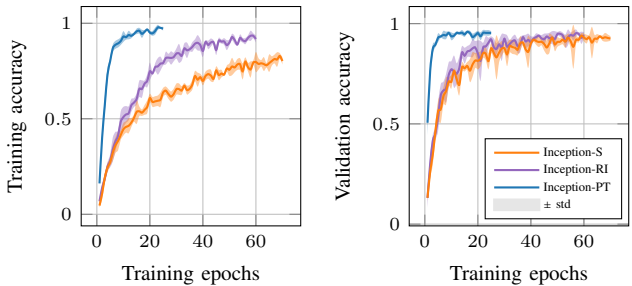
\begin{figure}[t]
    \vspace{0.5em}
    \centering
    \begin{subfigure}[t]{0.45\linewidth}
        \hspace{-1em}
        \tikzsetnextfilename{text_train_acc}
        \begin{tikzpicture}[font=\scriptsize]
            \begin{axis}[rewplot, ylabel = {\footnotesize Training accuracy}, legend style={font=\footnotesize}, width=4.5cm]
                \plotstdcom{pretrained_train_summary.csv}{pltBlue}{1}{PRE}{step}{mean}[std][1][1];
                \plotstdcom{random_weights_train_summary.csv}{pltPurple}{1}{RAND}{step}{mean}[std][1][1]; 
                \plotstdcom{small_inception_train_summary.csv}{pltOrange}{1}{SMALL}{step}{mean}[std][1][1];
                \legend{};
            \end{axis}
        \end{tikzpicture}
        \label{fig:texture-class-train}
        \vspace{-0.5em}
    \end{subfigure}
    \quad
    \begin{subfigure}[t]{0.45\linewidth}
        \hspace{-1em}
        \tikzsetnextfilename{text_val_acc}
        \begin{tikzpicture}[font=\scriptsize]
            \begin{axis}[rewplot, ylabel = {\footnotesize Validation accuracy}, legend style={font=\footnotesize}, width=4.5cm]
                \addlegendimage{std legend,fill=gray!20,draw=gray!20,mark=none}
                \addlegendentry{$\pm$ std}
                \plotstdcom{pretrained_val_summary.csv}{pltBlue}{1}{Inception-PT}{step}{mean}[std][1][1];
                \plotstdcom{random_weights_val_summary.csv}{pltPurple}{1}{Inception-RI}{step}{mean}[std][1][1];           
                \plotstdcom{small_inception_val_summary.csv}{pltOrange}{1}{Inception-S}{step}{mean}[std][1][1];
            \end{axis}
        \end{tikzpicture}
        \label{fig:texture-class-valid}
        \vspace{-1em}
    \end{subfigure}
    \caption{Training and validation accuracy of the Inception-v3 models (see~\cref{subsec:prob_classifier}) on the non-interactive 25-fabric classification task, averaged over five runs.
    Each model is trained until the validation accuracy converges.
    The final validation accuracies are $95.2\%$ for \ptModel, $94.2\%$ for \rawModel, and $92.6\%$ for \smallModel.}
    \vspace{-1em}
    \label{fig:imgclass_perf}
\end{figure}

Our experiments aim at identifying what components of the algorithmic architecture matter for active texture recognition with vision-based tactile sensors.
For that, we first compare the three probabilistic classifier architectures introduced in~\cref{subsec:prob_classifier} in a classical, non-interactive supervised learning setting on all 25 fabrics in our dataset.
Second, we evaluate the active sample selection strategies from \cref{subsec:strategy_intro}.
Third, we present a human study in which we investigate human exploration strategies on the same task in order to identify potential improvements to the robotic policies.
Fourth, we provide ablation studies of the effects of hyperparameters on a bigger \newdataset~\cite{yuan2018active}.

\subsection{Supervised Texture Classification}
\label{subsec:textclass}

In this experiment, the three models introduced in \cref{subsec:prob_classifier} are trained to distinguish all 25 fabrics of our dataset.
Since we are not interested in uncertainty quantification in this experiment, we set the dropout probability to 0\%.
For this experiment, we use 10 images per class for training and 20 images for validation. 
We see considerable differences within the performance of the three models. \cref{fig:imgclass_perf} shows that pre-training helps to learn to recognize the textures quickly and that \smallModel~needs more training time to reach the same performance as the other two models.
At the same time, the performance on the validation data is close to \rawModel, indicating that \smallModel~still generalizes well.

\subsection{Active Texture Recognition}
\label{subsec:active_results}

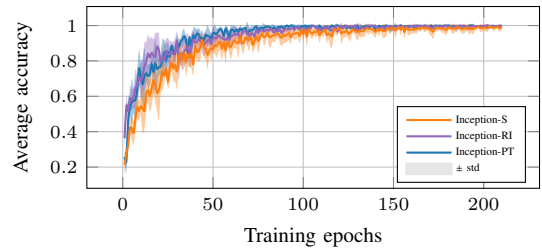
\begin{figure}[t]
    \vspace{0.5em}
    \hspace{-1em}
    \centering
    \begin{subfigure}[t]{0.8\linewidth}
        \tikzsetnextfilename{active_perf_models}
        \begin{tikzpicture}[font=\scriptsize]
            \begin{axis}[rewplot, ylabel = {\footnotesize Average accuracy}, height=4cm]
                \addlegendimage{std legend,fill=gray!20,draw=gray!20,mark=none}
                \addlegendentry{$\pm$ std}
                \plotstdcom{pretrained_acc_summary.csv}{pltBlue}{1}{Inception-PT}{step}{mean}[std][1][1];
                \plotstdcom{random_weights_acc_summary.csv}{pltPurple}{1}{Inception-RI}{step}{mean}[std][1][1]; 
                \plotstdcom{small_inception_acc_summary.csv}{pltOrange}{1}{Inception-S}{step}{mean}[std][1][1];
            \end{axis}
        \end{tikzpicture}
    \end{subfigure}
    \caption{Comparing the performance of the Inception-v3 models on the  active texture recognition task. Notably, the small Inception network \smallModel~performs as well as the larger \ptModel.}
    \label{fig:active_perf_models}
    \vspace{-0.5em}
\end{figure}

The objective of this experiment is to compare the three network architectures (\cref{subsec:prob_classifier}) and the four active sampling strategies (\cref{subsec:strategy_intro}).
Each model is trained for 210 epochs, 10 epochs after creating a baseline and then 10 more epochs after resampling in each of the 20 rounds. 
For all three models, we collect the results of five subsets of fabrics using the four different strategies and average the performance of each model. 
In \cref{fig:active_perf_models}, it can be seen that the models perform similarly, unlike in the non-active experiment in \cref{subsec:textclass}.
Especially the large advantage of \ptModel~seen in \cref{subsec:textclass} cannot be observed in this experiment.
We believe the reason why pre-training is not advantageous in this case is the retrospective addition of dropout layers, which the model was not trained for.
Considering that \smallModel~can solve the task on par with the other models while being substantially less computationally expensive, we use this model for our further comparisons. 

\begin{figure*}[t]
    \vspace{0.5em}
    \centering
    \begin{subfigure}[t]{0.32\linewidth}
        \centering
        \tikzsetnextfilename{active_acc_no_outlier}
        \begin{tikzpicture}[font=\scriptsize]
            \begin{axis}[rewplot, ylabel = {\footnotesize Average accuracy}, legend style={font=\footnotesize}]
                \addlegendimage{std legend,fill=gray!20,draw=gray!20,mark=none}
                \addlegendentry{$\pm$ std}
                \plotstdcom{var_acc_summary.csv}{pltOrange}{1}{VAR}{step}{mean}[std][1][1];
                \plotstdcom{entr_acc_summary.csv}{pltGreen}{1}{ENTR}{step}{mean}[std][1][1]; 
                \plotstdcom{random_acc_summary.csv}{pltLightBlue}{1}{RAND}{step}{mean}[std][1][1];
                \plotstdcom{YOTO_acc_summary.csv}{pltPink}{1}{YOTO}{step}{mean}[std][1][1];
            \end{axis}
        \end{tikzpicture}
    \end{subfigure}
    \begin{subfigure}[t]{0.32\linewidth}
        \centering
        \tikzsetnextfilename{active_var_no_outlier}
        \begin{tikzpicture}[font=\scriptsize]
            \begin{axis}[rewplot, ylabel = {\footnotesize Average variance}, xlabel={\footnotesize Rounds}, legend style={font=\footnotesize}]
                \plotstdcom{var_var_summary.csv}{pltOrange}{1}{VAR}{step}{mean}[std][1][1];
                \plotstdcom{entr_var_summary.csv}{pltGreen}{1}{ENTR}{step}{mean}[std][1][1]; 
                \plotstdcom{random_var_summary.csv}{pltLightBlue}{1}{RAND}{step}{mean}[std][1][1];
                \plotstdcom{YOTO_var_summary.csv}{pltPink}{1}{YOTO}{step}{mean}[std][1][1];
                \legend{};
            \end{axis}
        \end{tikzpicture}
    \end{subfigure}
    \begin{subfigure}[t]{0.32\linewidth}
        \centering
        \tikzsetnextfilename{active_entr_no_outlier}
        \begin{tikzpicture}[font=\scriptsize]
            \begin{axis}[rewplot, ylabel = {\footnotesize Entropy}, xlabel={\footnotesize Rounds}, legend style={font=\footnotesize}]
                \addlegendimage{std legend,fill=gray!20,draw=gray!20,mark=none}
                \addlegendentry{$\pm$ std}
                \plotstdcom{var_entr_summary.csv}{pltOrange}{1}{VAR}{step}{mean}[std][1][1];
                \plotstdcom{entr_entr_summary.csv}{pltGreen}{1}{ENTR}{step}{mean}[std][1][1]; 
                \plotstdcom{random_entr_summary.csv}{pltLightBlue}{1}{RAND}{step}{mean}[std][1][1];
                \plotstdcom{YOTO_entr_summary.csv}{pltPink}{1}{YOTO}{step}{mean}[std][1][1];
                \legend{};
            \end{axis}
        \end{tikzpicture}
    \end{subfigure}
    \caption{Comparison of the exploration strategies on the \emph{tactile active texture recognition} task.
    Average prediction accuracy, average variance, and entropy of the predictions are shown.
    \smallModel~is used in all experiments.
    The \emph{Variance} strategy achieves the highest accuracy, closely followed by \emph{Entropy} and \emph{Random}.
    Interestingly, the \emph{Variance} strategy leads to a faster entropy decrease even than \emph{Entropy} (rightmost plot).
    }
    \label{fig:active_perf}
    \vspace{-1em}
\end{figure*}
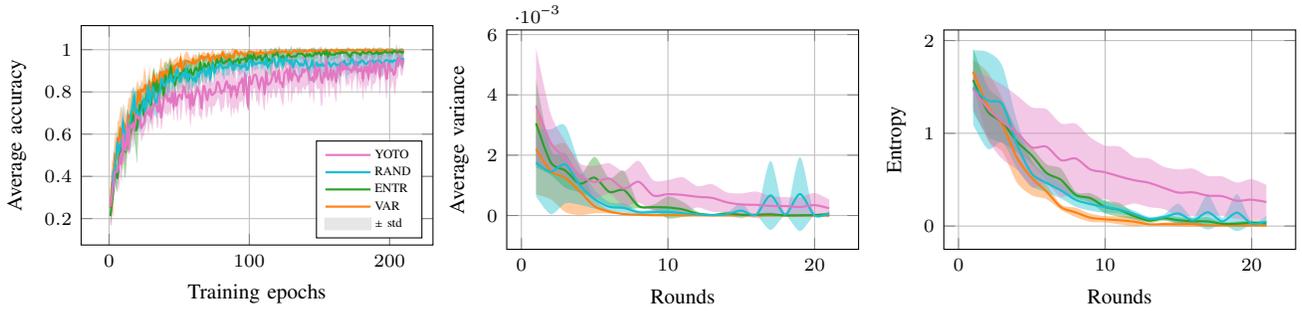

In \cref{fig:active_perf}, the influence of the different strategies on the performance of \smallModel~is shown.
When we run the experiment for 20 rounds, sampling offers an advantage, as \emph{YOTO} has the lowest performance on the training data. 
On average, the model performs best using \emph{Variance}, closely followed by \emph{Entropy} and \emph{Random}.
However, the active  sampling strategies generally have a lower predictive uncertainty than \emph{YOTO}. 
While \emph{YOTO} performs quite well on the training data after 20 rounds, its accuracy in predicting the label of the reference object of each trial is only 80\% on average.

\subsection{Human Study}
\label{subsec:human_experiment}

In order to find out how well the proposed tactile active recognition method performs compared to humans, we carry out an experiment with ten human participants.
Ideally, we would like to see whether human exploration strategies can be characterized using information-theoretic metrics, and whether insights from humans can be transferred to the robots.

During a trial, the participants are blindfolded and can only use the tips of either their index or middle fingers to explore the fabrics (see \cref{subfig:human_setup}).
All participants have passed the two-point discrimination test with a point distance of 2mm, confirming that their tactile perception capabilities are not impaired~\cite{shooter2005use}.
They indicate their response to the experimenter by resting their finger on the chosen fabric and by verbally confirming their selection. 
No feedback on the participant's performance is provided during the experiment.

\begin{figure}[t]
    \centering
    \includegraphics[width=0.8\linewidth, height=3.7cm]{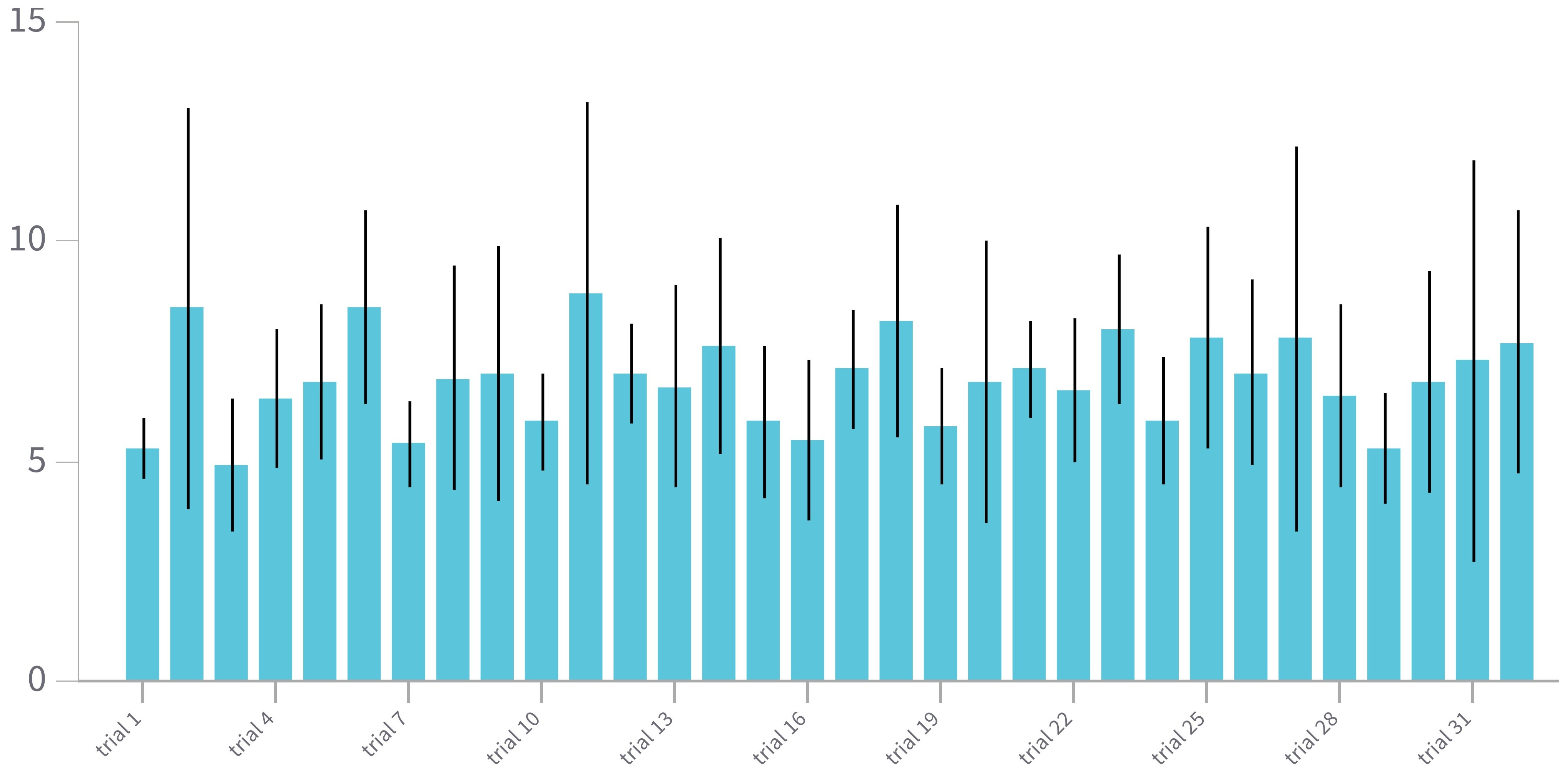}
    \caption{The average number of touches made by the human participants in each trial. Error bars denote the standard error. Humans do $5$--$9$ touches (i.e., $1$--$5$ rounds) in each trial before giving the final response. Some harder trials with similar objects show high variance.}
    \label{fig:nb_touches}
\end{figure}

\begin{table}[t]
\centering
\begin{tabular}{|c|c|c|c|c|} 
 \hline
 Humans & \emph{Variance} & \emph{Entropy} & \emph{Random} & \emph{YOTO} \\ 
 \hline
 \makecell{$66.88\%$ \\ $\pm 16.93\%$} & \makecell{$\textbf{90.00}\%$ \\ $\pm 15.24\%$} & \makecell{$88.13\%$ \\ $\pm 14.24\%$} & \makecell{$89.38\%$ \\ $\pm 14.35\%$}  & \makecell{$80.63\%$ \\ $\pm 22.42\%$} \\ 
 \hline
\end{tabular}
\caption{Comparison of the final accuracies achieved by the different exploration strategies.
\emph{Humans} denotes the average human performance.
Notably, all robotic strategies are superior to humans, showing that the vision-based tactile sensor alone provides an advantage over the human touch in this task.
The type of the exploration strategy, however, seems to play a minor role, since \emph{Variance}, \emph{Entropy}, and \emph{Random} all achieve about $90\%$ accuracy.
}
\label{table:experiment_acc}
\vspace{-1em}
\end{table}

We select $8$ out of $25$ fabrics which were mostly confused by the neural network and we use each fabric as the reference object four times, with the corresponding comparison object placed in each of the four possible locations once, resulting in 32 trials in total.
To analyze the number and time of revisits per object, we record videos of each participant's hand movements.
Each time a participant switches between two objects is counted as a new revisit.
The data of this experiment and the code for analysis are available online.\footnote{Our study data
\url{https://drive.google.com/drive/folders/1S_2PLKV-Ap2tifV1gvMfCYjoaNyQaF9z?usp=sharing}}

Figure~\ref{fig:nb_touches} shows the average number of revisits before giving a response in each trial, ranging from five to nine revisits needed per trial. 
In~\cref{table:experiment_acc}, we compare the prediction accuracy of the human participants and the robot.
To ensure fairness, in each trial, we only allow the robot to use the same number of touches that humans used (\cref{fig:nb_touches}).
The low prediction accuracy of $66.88\%$ achieved by the humans indicates that the task is quite non-trivial, and it provides a reference point for the $90\%$ accuracy achieved using the vision-based tactile sensor.
In the next section, we take a closer look at the exploration strategies employed by the humans in comparison to the information-theoretic strategies.

\subsection{Behavior Comparison Between Participants and the Robot}
\label{subsec:strategy_comparison}

\begin{figure}[t]
    \centering
    \begin{subfigure}{0.45\linewidth}
    \includegraphics[height=2.5cm, width=0.87\linewidth]{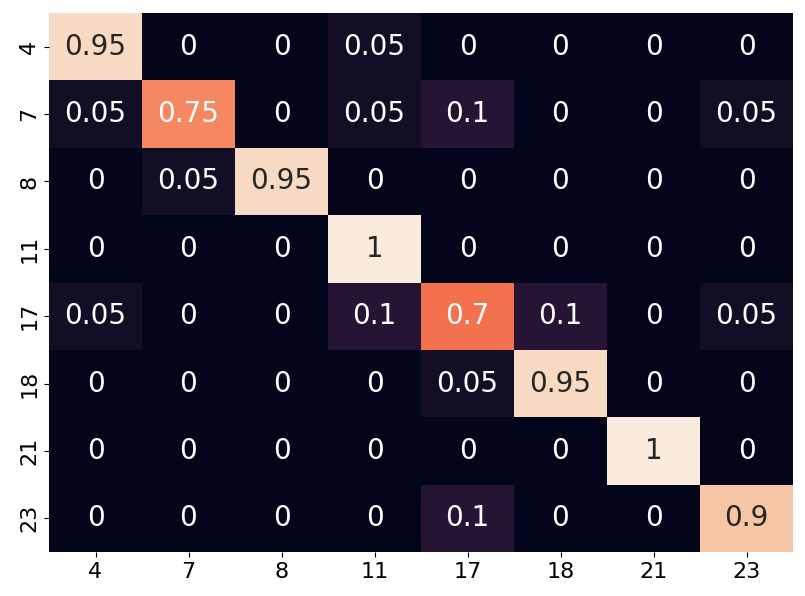}
    \caption{\emph{Variance} strategy}
    \end{subfigure}
    \begin{subfigure}{0.45\linewidth}
    \includegraphics[height=2.5cm, width=0.87\linewidth]{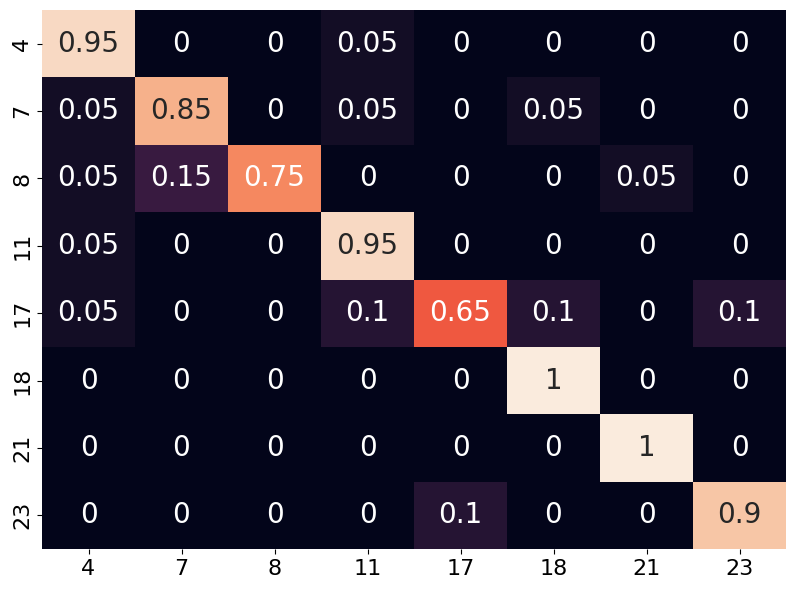}
    \caption{\emph{Entropy} strategy}
    \end{subfigure}
    \begin{subfigure}{0.45\linewidth}
    \includegraphics[height=2.5cm, width=0.87\linewidth]{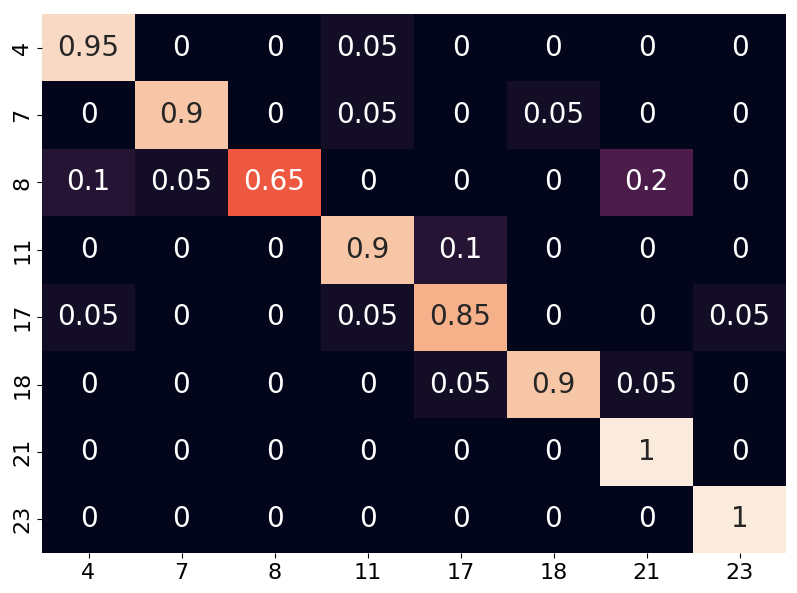}
    \caption{\emph{Random} strategy}
    \end{subfigure}
    \begin{subfigure}{0.45\linewidth}
    \includegraphics[height=2.5cm, width=\linewidth]{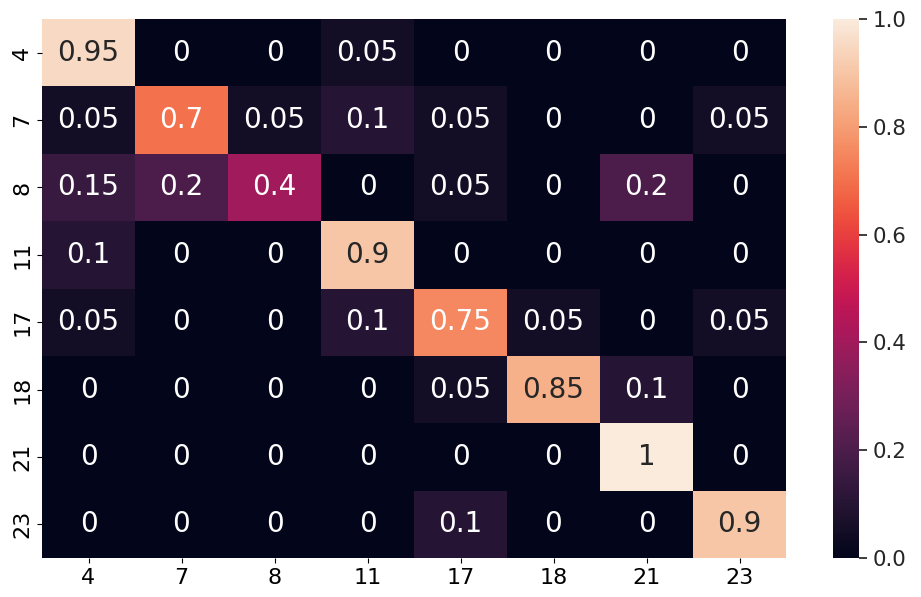}
    \caption{\emph{YOTO} strategy}
    \end{subfigure}
    \caption{Confusion matrices for the $8$ fabrics included in the human study. Some fabrics are consistently misclassified by all strategies, e.g., $17$ and $7$, while others are always classified correctly, e.g., $21$.}
    \label{fig:confusion}
    \vspace{-1em}
\end{figure}

\begin{figure}[t]
    \centering
    \begin{subfigure}{0.42\linewidth}
        \centering
        \includegraphics[height=2.8cm, width=\linewidth]{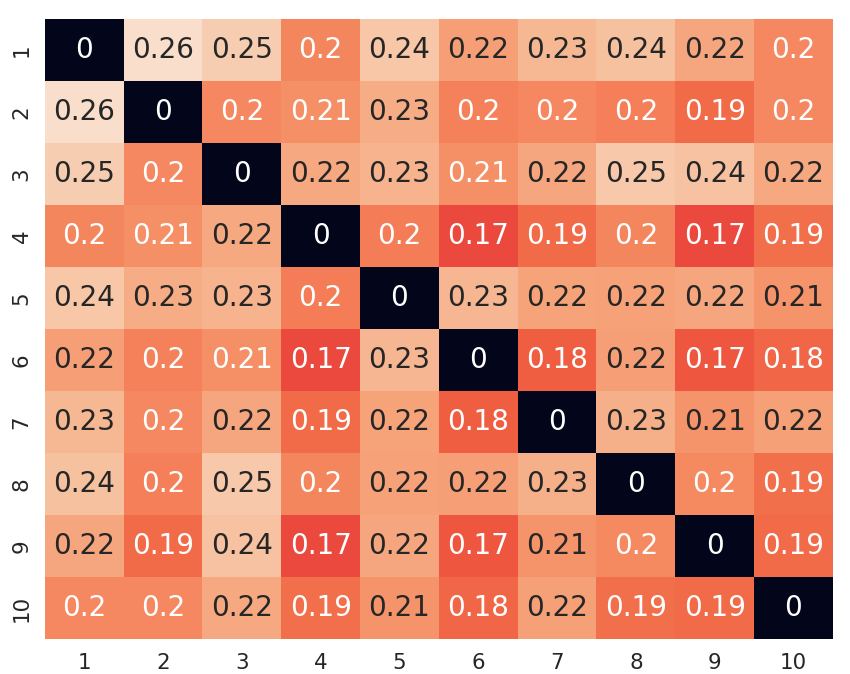}
        \caption{inter-participant}
        \label{subfig:js_part}
    \end{subfigure}
    \begin{subfigure}{0.42\linewidth}
        \centering
        \includegraphics[height=2.8cm, width=\linewidth]{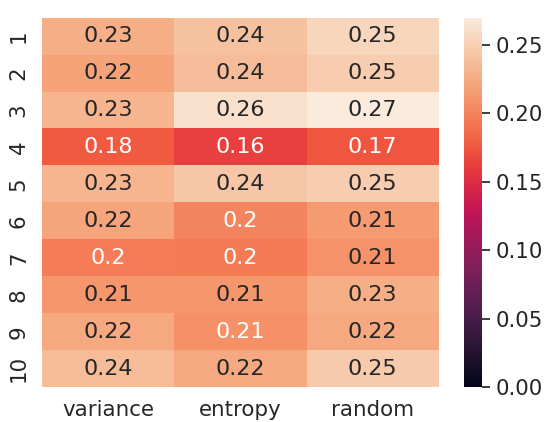}
        \caption{participant-robot}
        \label{subfig:js_part-robot}
    \end{subfigure}
    \caption{Comparing the exploration strategies among participants and against the information-theoretic strategies.
    The numbers indicate the Jensen-Shannon divergence between the distributions of time spent over objects, averaged over trials.
    The inter-subject variability is comparable to the subject-robot variability, therefore no uniform judgement about what strategy all humans use can be made.
    Instead, each participant seems to follow a personal exploration strategy.
    }
    \label{fig:js}
    \vspace{-1.5em}
\end{figure}

To gain insight into the difficulty of differentiating the fabrics, we plot the confusion matrices in \cref{fig:confusion}.
Some fabrics have close to $1$ recognition accuracy, whereas others are misclassified more often.
Notably, the confusion matrix of \emph{YOTO} has the lowest values on the diagonal, in accordance with the results in~\cref{table:experiment_acc}.
For humans, an average confusion matrix is not very informative since no single fabric was inherently harder to recognize for all participants, i.e., the inter-participant variance was relatively high.

To compare the exploration strategies, we formalize the problem by normalizing the time spent by the human participants on each object per trial, to get a distribution of relative times per fabric.
This gives us a distribution of time spent over objects, and subsequently we can compute a distance between these distributions to judge how close they are.
We employ the symmetric Jensen-Shannon divergence.
Thus, we can compare both human and robotic strategies to each other at least in this restricted sense.
\emph{YOTO} is excluded from this comparison as it performs no exploration. 

The Jensen-Shannon distance takes values in the range $[0, 1]$, with lower values indicating greater similarity between strategies.
The resulting distances of comparing the robotic strategies to each other are 0.14 between \emph{Variance} and \emph{Entropy}, 0.12 between \emph{Entropy} and \emph{Random}, and 0.16 between \emph{Variance} and \emph{Random}.
Thus, according to the Jensen-Shannon divergence, the two uncertainty-based strategies \emph{Variance} and \emph{Entropy} are not the most similar, and \emph{Entropy} produces exploratory behavior that is more similar to \emph{Random} than to \emph{Variance} on average. While the distances between the robot strategies are in the range 0.12--0.16, the inter-participant~(\cref{subfig:js_part}) and the participant-robot~(\cref{subfig:js_part-robot}) distances are around 0.2 and higher.

If we average the distances of each participant, we get a mean Jensen-Shannon distance of 0.219 for the variance strategy, 0.218 for the entropy strategy, and 0.231 for the random sampling strategy, meaning that the uncertainty-based strategies are on average slightly more similar to human exploration under the Jensen-Shannon distance.
On the other hand, we again observe a high variance between trials. There are some trials where the same two participants follow a very similar exploration strategy and others where they choose different approaches with a higher Jensen-Shannon distance.

Finally, we observe a similar pattern when it comes to the \emph{Variance} strategy of the robot and the human participants. 
In 63.75\% of the trials, the participants touched that object most often which they predicted to be the reference object. 
With \emph{Variance}, we get 56.25\%, and with \emph{Entropy}, only 32.5\% of the trials where the predicted object is touched most often.

\subsection{Ablation Study on the Active Clothing Perception Dataset}

\begin{table}[t]
\vspace{0.25em}
\centering
\begin{tabular}{|c|c|c|c|c|} 
 \hline
 & \emph{Variance} & \emph{Entropy} & \emph{Random} & \emph{YOTO} \\ 
 \hline
 DA, DR=0.25 & \makecell{$68.13\%$ \\ $\pm 46.8\%$} & \makecell{$78.44\%$ \\ $\pm 40.81\%$} & \makecell{$81.87\%$ \\ $\pm 37.02\%$} & \makecell{$46.88\%$ \\ $\pm 50.7\%$} \\ 
 \hline
 No DA & \makecell{$50\%$ \\ $\pm 50.8\%$} & \makecell{$59.22\%$ \\ $\pm 48.33\%$} & \makecell{$62.66\%$ \\ $\pm 47.38\%$} & \makecell{$49.22\%$ \\ $\pm 50.19\%$} \\ 
  \hline
 DR=0.5 & \makecell{$55.94\%$ \\ $\pm 50.15\%$} & \makecell{$80.94\%$ \\ $\pm 37.01\%$} & \makecell{$79.06\%$ \\ $\pm 39.54\%$} & \makecell{$49.06\%$ \\ $\pm 50.12\%$} \\ 
 \hline
 DR=0.15 & \makecell{$70.94\%$ \\ $\pm 45.25\%$} & \makecell{$80.31\%$ \\ $\pm 39.55\%$} & \makecell{$79.69\%$ \\ $\pm 36.23\%$} & \makecell{$47.19\%$ \\ $\pm 50.43\%$} \\ 
 \hline
 DR=0.05 & \makecell{$68.75\%$ \\ $\pm 47.09\%$} & \makecell{$84.38\%$ \\ $\pm 36.89\%$} & \makecell{$86.56\%$ \\ $\pm 33.47\%$} & \makecell{$50\%$ \\ $\pm 49.25\%$} \\ 
 \hline
\end{tabular}
\caption{Ablations on the influence of Data Augmentation (DA) and Dropout Rate (DR) on the final prediction accuracy of different exploration strategies, evaluated on the \newdataset~\cite{yuan2018active}.
Adding data augmentation boosts the performance by $20\%$ for all  strategies except \emph{YOTO}.
Decreasing the dropout rate generally improves the results, but to a lesser degree.
}
\vspace{-1.25em}
\label{table:experiment_acc_newdata}
\end{table}

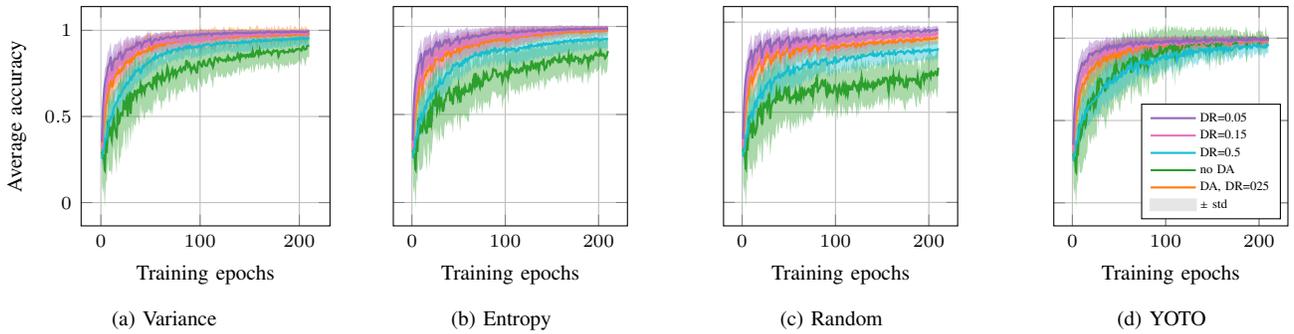
\begin{figure*}[t]
    \vspace{0.5em}
    \centering
    \begin{subfigure}[t]{0.25\linewidth}
        \centering
        \tikzsetnextfilename{active_acc_var.png}
        \begin{tikzpicture}[font=\scriptsize]
            \begin{axis}[width=0.5\linewidth, rewplot, ylabel = {\footnotesize Average accuracy}, legend style={font=\footnotesize}]
                \plotstdcom{aug_var_acc_summary.csv}{pltOrange}{1}{1}{step}{mean}[std][1][1];
                \plotstdcom{noaug_var_acc_summary.csv}{pltGreen}{1}{2}{step}{mean}[std][1][1]; 
                \plotstdcom{drop05_var_acc_summary.csv}{pltLightBlue}{1}{3}{step}{mean}[std][1][1];
                \plotstdcom{drop015_var_acc_summary.csv}{pltPink}{1}{4}{step}{mean}[std][1][1];
                \plotstdcom{drop005_var_acc_summary.csv}{pltPurple}{1}{5}{step}{mean}[std][1][1];
                \legend{};
            \end{axis}
        \end{tikzpicture}
        \caption{Variance}
    \end{subfigure}
    \begin{subfigure}[t]{0.24\linewidth}
        \centering
        \tikzsetnextfilename{active_acc_entr.png}
        \begin{tikzpicture}[font=\scriptsize]
            \begin{axis}[rewplot, legend style={font=\footnotesize}, yticklabels={,,}]
                \plotstdcom{aug_entr_acc_summary.csv}{pltOrange}{1}{1}{step}{mean}[std][1][1];
                \plotstdcom{noaug_entr_acc_summary.csv}{pltGreen}{1}{2}{step}{mean}[std][1][1]; 
                \plotstdcom{drop05_entr_acc_summary.csv}{pltLightBlue}{1}{3}{step}{mean}[std][1][1];
                \plotstdcom{drop015_entr_acc_summary.csv}{pltPink}{1}{4}{step}{mean}[std][1][1];
                \plotstdcom{drop005_entr_acc_summary.csv}{pltPurple}{1}{5}{step}{mean}[std][1][1];
                \legend{};
            \end{axis}
        \end{tikzpicture}
        \caption{Entropy}
    \end{subfigure}
    \begin{subfigure}[t]{0.24\linewidth}
        \centering
        \tikzsetnextfilename{active_acc_random.png}
        \begin{tikzpicture}[font=\scriptsize]
            \begin{axis}[rewplot, legend style={font=\footnotesize}, yticklabels={,,}]
                \plotstdcom{aug_random_acc_summary.csv}{pltOrange}{1}{1}{step}{mean}[std][1][1];
                \plotstdcom{noaug_random_acc_summary.csv}{pltGreen}{1}{2}{step}{mean}[std][1][1]; 
                \plotstdcom{drop05_random_acc_summary.csv}{pltLightBlue}{1}{3}{step}{mean}[std][1][1];
                \plotstdcom{drop015_random_acc_summary.csv}{pltPink}{1}{4}{step}{mean}[std][1][1];
                \plotstdcom{drop005_random_acc_summary.csv}{pltPurple}{1}{5}{step}{mean}[std][1][1];
                \legend{};
            \end{axis}
        \end{tikzpicture}
        \caption{Random}
    \end{subfigure}
    \begin{subfigure}[t]{0.24\linewidth}
        \centering
        \tikzsetnextfilename{active_acc_yoto.png}
        \begin{tikzpicture}[font=\scriptsize]
            \begin{axis}[rewplot, legend style={font=\footnotesize}, yticklabels={,,}]
                \addlegendimage{std legend,fill=gray!20,draw=gray!20,mark=none}
                \addlegendentry{$\pm$ std}
                \plotstdcom{aug_YOTO_acc_summary.csv}{pltOrange}{1}{DA, DR=025}{step}{mean}[std][1][1];
                \plotstdcom{noaug_YOTO_acc_summary.csv}{pltGreen}{1}{no DA}{step}{mean}[std][1][1]; 
                \plotstdcom{drop05_YOTO_acc_summary.csv}{pltLightBlue}{1}{DR=0.5}{step}{mean}[std][1][1];
                \plotstdcom{drop015_YOTO_acc_summary.csv}{pltPink}{1}{DR=0.15}{step}{mean}[std][1][1];
                \plotstdcom{drop005_YOTO_acc_summary.csv}{pltPurple}{1}{DR=0.05}{step}{mean}[std][1][1];
            \end{axis}
        \end{tikzpicture}
        \caption{YOTO}
    \end{subfigure}
    \caption{Ablations on the influence of Data Augmentation (DA) and Dropout Rate (DR) on the prediction accuracy during training for different exploration strategies, evaluated on the \newdataset~\cite{yuan2018active}.
    Adding data augmentation significantly improves the performance for all  strategies except \emph{YOTO}.
    Decreasing the dropout rate generally improves the results, but to a lesser degree.}
    \label{fig:active_perf_newdata}
    \vspace{-1em}
\end{figure*}

In the ablation study, we investigate the role of other hyperparameters and design choices on the performance of the considered tactile active texture recognition algorithm.
To make sure that the results are not specific to our dataset, we perform the ablations on the images from the~\newdataset~\cite{yuan2018active}.
To make the setup comparable to our experiments, we randomly select 8 of the fabrics and create 32 trials, with 4 fabrics each.

Table~\ref{table:experiment_acc_newdata} and \cref{fig:active_perf_newdata} show the results of the ablation studies on the Dropout Rate (DR) and Data Augmentation (DA).
In general, the dataset~\cite{yuan2018active} contains more variable data compared to ours, because the data was collected by autonomously grasping real clothes at wrinkle locations.
Furthermore, a prior version of the GelSight sensor was used, that exhibits a higher variance in the light distribution upon contact.
For these reasons, the variance in the performance of our method is higher on this dataset (\cref{table:experiment_acc_newdata}) compared to ours (\cref{table:experiment_acc}).

In the first two rows in \cref{table:experiment_acc_newdata}, we compare the final average test accuracies with and without DA, at a fixed $\mathrm{DR}=0.25$.
For all methods except \emph{YOTO}, data augmentation adds about $20\%$ to the accuracy.
This confirms our observation that data augmentation is essential for the good performance of the CNN-based model.

In the bottom three rows in \cref{table:experiment_acc_newdata}, we compare different dropout rates.
The general trend is that the smaller DR results in a higher average accuracy, though the improvement is minor in the range $\mathrm{DR}\leq0.25$.
Therefore, in our main experiments, we used $0.25$ as it performed sufficiently well in all tests.

In~\cref{fig:active_perf_newdata} we observe the same trends during training as in~\cref{table:experiment_acc_newdata}.
Namely, removing data augmentation leads to a significant drop in performance, and decreasing the dropout rate provides an improvement to the accuracy of all strategies.

\section{Discussion \& Conclusion}
\label{sec:conclusion}
We have investigated the performance of a Bayesian approach to active sampling for fabric texture recognition with vision-based tactile sensors using variance and entropy criteria.
We performed ablation studies on different model architectures and hyperparameters, and we identified which choices have the largest impact on the recognition accuracy.

First, we found that an ImageNet-pretrained Inception-v3 network allows for a significantly faster training of a 25-class classifier on our dataset of denim and cotton fabrics (see \cref{fig:imgclass_perf}), achieving $95.2\%$ accuracy after $20$ epochs of training.
However, on our main $4$-class recognition task, where the network needs to adapt very quickly with only a handful of training samples, we found that even a much smaller model \smallModel~performs similarly, while being more computationally efficient (see \cref{fig:active_perf_models}).
Therefore, we conclude that big pretrained networks are not necessary for  few-shot recognition tasks with vision-based tactile sensors.

Second, we investigated the importance of the exploration strategy on the recognition accuracy, and we did not find a significant difference between the strategies that sample the objects with highest predictive variance or entropy contribution.
Furthermore, even a random sampling strategy has shown similar performance, which suggests that the texture recognition task is relatively straightforward for the vision-based tactile sensors such as GelSight Mini, despite the fact that humans only achieved $66.88\%$ average accuracy on this task.
These results are in agreement with the work on active classification of material roughness~\cite{amini2020uncertainty}, where the algorithm using vision-based tactile sensors was shown to achieve significantly higher accuracy than human participants.

Third, we performed a human study with the goal of quantifying the human performance on the texture recognition tasks, and we performed analysis of exploratory behaviors that humans employ in order to compare them to the behavior of our exploration strategies.
Apart from confirming that the task is quite hard for humans, we found out that there is a significant variability among the participants with regards to the exploration strategy, as evidenced by our analysis in \cref{fig:js}.
Moreover, the inter-participant variability was found to be similar to the participant-robot variability, meaning that there is no universal exploration strategy that all participants have followed.
Nevertheless, on average, human exploration behavior was closer to the information-theoretic strategies, \emph{Variance} and \emph{Entropy}, than to random exploration.

Fourth, we reported the results of ablation studies on the effect of data augmentation and dropout rate on the model performance.
Most importantly, we found data augmentation to significantly improve the performance of all exploration strategies (see \cref{table:experiment_acc_newdata}) by almost $20\%$ on average.
The dropout rate, on the other hand, had a relatively smaller influence, well within the standard error range.
Combined with our observations about the influence of exploration strategies, this result allows us to conclude that the quality of the data and data augmentation, together with the network architecture, play a more significant role in improving the performance compared to the choice of the exploration strategy.

\subsubsection*{Limitations}
\label{sec:limitations}
Comparing human and robotic tactile perception in our study is limited due to the different nature of the sensors.
The vision-based tactile sensor achieving a higher performance on the texture recognition task could in principle be attributed to using a different sensing modality rather than to a better representation or sample selection strategy.
However, this concern was partially addressed in~\cite{amini2020uncertainty}, where human performance using touch was compared to using GelSight images for material roughness classification.
Interestingly, they found that humans are much better at classification using their sense of touch rather than vision.

Our results furthre suggest that the choice of which object to touch next in the fabric recognition task may depend on the nature of the internal representation, and not on the sampling strategy. It is, however, not straightforward to compare internal representations of humans and robots, especially given that there seems to be no `average' human representation, i.e., humans differ substantially in terms of which fabrics they confuse and which exploration strategy they follow.

\subsubsection*{Outlook}
\label{outlook}
While we evaluated ImageNet pre-trained networks on texture recognition, it would be of interest to develop a ``tactile ImageNet'' dataset and a network pre-trained only on textures. 
Such a network would potentially further improve the active texture recognition performance.
Tackling more challenging tasks, such as contour/shape exploration and object pose estimation would provide further insights into active tactile sensing, as well as integrating multiple sensing modalities, such as touch, vision, and proprioception.
On the human side, a study on the representations of object properties that humans utilize would be especially relevant.

\bibliographystyle{IEEEtran}
\bibliography{IEEEabrv,lit}

\end{document}